\documentclass[11pt,twoside,a4paper]{IEEEtran}
\usepackage{subfigure}
\usepackage{amsmath}
\usepackage{amssymb}
\usepackage{setspace}
\usepackage{geometry} 
\usepackage{graphicx}
\usepackage{multirow}
\usepackage{subfig}
\begin{document}

\title{Deep Neural Networks for automatic extraction of features in time series satellite images}

\author{
 Gael Kamdem De Teyou, Yuliya Tarabalka, Isabelle Manighetti, Rafael Almar, Sebastien  Tripodi}

\maketitle

\abstract
{

Many earth observation programs such as Landsat, Sentinel, SPOT, and Pleiades produce huge volume of medium to high resolution multi-spectral images every day that can be organized in time series. In this work, we exploit both temporal and spatial information provided by these images to generate land cover maps. For this purpose, we combine a fully convolutional neural network with a convolutional long short-term memory. Implementation details of the proposed spatio-temporal neural network architecture are provided. Experimental results show that the temporal information provided by time series images allows increasing the accuracy of land cover classification, thus producing up-to-date maps that can help in identifying changes on earth.

}

keywords: Deep Learning, U-Net, ConvLSTM, FCN, Time Series, Image Segmentation, Sentinel, remote sensing



\section{INTRODUCTION}\label{MANUSCRIPT}
Several places on earth are threatened by the ongoing climatic and anthropogenic global changes: with a low topography, a weak geological substratum, a poor fresh water supply, and a dense and rapidly expanding population, some areas are highly vulnerable to current sea level rise, extreme climatic phenomena, erosion, and modifications of the ecosystems and resources. Many earth observation programs such as Landsat, Sentinel, SPOT and Pleiades produce huge volume of medium to high resolution multispectral images every day that can be organized in time series and used to produce accurate and up-to-date land cover maps that can monitor environmental changes at different places and time ranges.

Land cover mapping is a semantic segmentation problem: each pixel in a satellite image must be classified into one of the land cover classes of interest. These classes describe the surface of the earth and are typically broad categories such as "water", "roads", "low vegetation", "forest", "building", etc. Throughout the years of research, a wide family of methods have been proposed,  ranging from the classification of individual pixels with machine learning techniques,  to the incorporation of higher-level information such as shape features. For this task, supervised machine learning algorithms have shown their potential, especially traditional algorithms such as Random Forests (RFs) and Support Vector Machines (SVMs). For example, in (Taati et al., 2014), a land use classification using SVM and maximum likelihood classifier (MLC) for Landsat images is presented. The article (Das et al., 2019) highlights the use of binary logistic regression for land-use land-cover (LULC) classification with Sentinel-2 multispectral images. In (Thanh Noi, Kappas, 2018), the authors provide a comparison of Random Forest, k-Nearest Neighbor, and SVM classifiers for land cover classification using Sentinel-2 imagery. 

In the past few years, Deep learning (DL),  a class of machine learning algorithms that uses multiple layers to progressively extract higher level features from complex data, has gained attention. DL made tremendous progress in the field of computer vision, with Artificial Neural Networks (ANNs) repeatedly pushing the frontier of visual recognition technology. Deep learning architectures, especially Fully Convolutional Networks (FCNs) show a great potential for application to various remote sensing problems such as land cover mapping. A FCN uses a convolutional neural network to transform image pixels to pixel categories so that the predictions have a one-to-one correspondence with input image in spatial dimension (Long et al., 2015). In (Priit, Innar, 2020) and (Syrris et al., 2019), U-Net a FCN that was developed for biomedical image segmentation, is used to produce land cover classification mapping based on Sentinel-2 images. In (Amina Ben Hamida et al., 2017) and (Nicolas Audebert, 2016), the authors modified SegNet, a deep convolutional encoder-decoder architecture for semantic pixel-wise labelling, for land cover classification from satellite images. In (Yao et al., 2019), land use classification based on a deep convolutional neural network reducing the loss of spatial features is proposed. The authors also tested  other FCN such as U-net, SegNet, and  Deeplab. More generally,  FCNs have shown promising performances for land cover mapping from satellite images (G. Sumbul et al., 2019). 

However, even if FCNs deal very  well with spatial representation of features within an image, they are unable to learn the additional information provided by the the multi-temporal structure of time series images, which can improve the land cover classification accuracy and efficiency. In addition, when applied on a single image, they lead systematically to a variance, that depends on the day when the image was acquired. To capture the temporal dependency of images, Recurrent Neural Networks (RNN), a class of deep learning architectures where connections between nodes form a directed graph along a temporal sequence, can be used. For example, (G. Sumbul, B. Demir, 2019) present a novel multi-attention driven system that jointly exploits Convolutional Neural Network (CNN) and RNNs in the context of multi-label remote sensing image classification. The article (Stoian et al., 2019) proposes a framework for working with Sentinel-2 L2A time-series image data, and an adaptation of the U-Net model for dealing with sparse annotation data while maintaining high resolution output. In (C. Pelletier et al., 2019), an analysis of RNN and Temporal Convolutional Neural Networks (TempCNNs) for the classification of Sentinel-2 image time series is provided. In this work, we exploit both temporal and spatial information provided by multi-temporal Sentinel-2 images to generate accurate and up-to-date land cover maps. For this purpose, we combine a U-Net with a RNN. The remainder of this paper is organized as follows. In Section \ref{methodology}, we present our methodology, including the design of the U-Net and RNN. In Section \ref{dataset}, we present the dataset we built and that was one of the key points of this work. Finally, experimental results are presented in Section \ref{results}.

\section{METHODOLOGY}
\label{methodology}
\begin{figure}[htp]
    \centering
    \includegraphics[width=6cm]{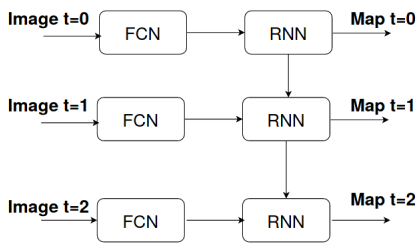}
    \caption{Architecture of the model}
    \label{architecture}
\end{figure}

An overview of our pipeline is shown on Fig. \ref{architecture}. It is a hybrid architecture combining a FCN and a RNN. The FCN captures the spatial representation of features in the image while the RNN learns the temporal variations of these features. In the next subsections, we discuss the design of each building box of the diagram of Fig. \ref{architecture}.

\subsection{FCN architecture}
\begin{figure}[htp]
    \centering
    \includegraphics[width=8.5cm]{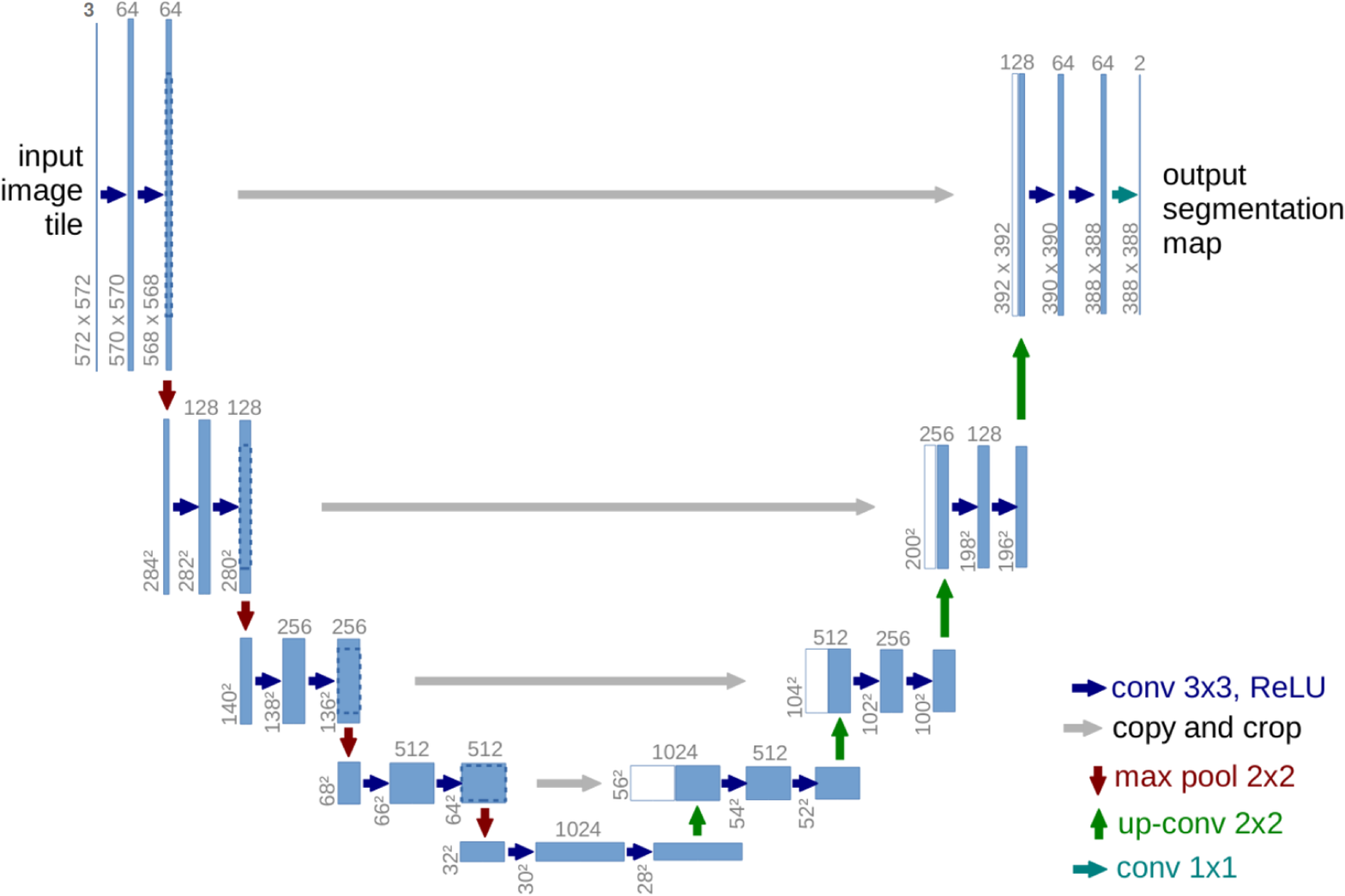}
    \caption{U-net architecture (example for 32x32 pixels in the lowest resolution). Each blue box corresponds to a multi-channel feature map. The number of channels is denoted
on top of the box. The x-y-size is provided at the lower left edge of the box. White
boxes represent copied feature maps. The arrows denote the different operations.}
    \label{unet_architecture}
\end{figure}

The architecture of our FCN is based on the U-Net (O. Ronneberger et al., 2015) and is depicted on Fig. \ref{unet_architecture}. It is a symmetric encoder-decoder structure consisting of  a contracting branch (left side) that captures the context and an expanding branch (right side) that enables precise localization for the segmentation masks. The contracting path consists of the repeated application of two 3 $\times$ 3 convolutions, each followed by a ReLu and a $2 \times 2$ max pooling for downsampling. At each downsampling step, the number of feature channels is doubled. Every step in the expansive path consists of a transpose convolution that halves the number of feature channels, a concatenation with the corresponding feature map from the contracting path and one 3x3 convolution. We used this architecture with only two modifications. First we used half as many filters at each layer (see Fig. \ref{unet_architecture}). Therefore, for example, we used 32 filters instead of 64 in the first-level convolutional layers, 64 filters instead of 128 filters in the second-level layers, etc. Second we inserted batch normalization after each convolutional layer to speed up convergence. 

This FCN enables us to generate a Probability Map (PMap) from an input image. The PMap is an image of the same size as the input but with $l_c$ channels, where $l_c$ is the number of land cover classes. For each pixel, it gives the probability to belong to a particular land cover class.

\subsection{RNN architecture}
\begin{figure}
\centering
\subfigure[April 4, 2020]{\includegraphics[width=6cm]{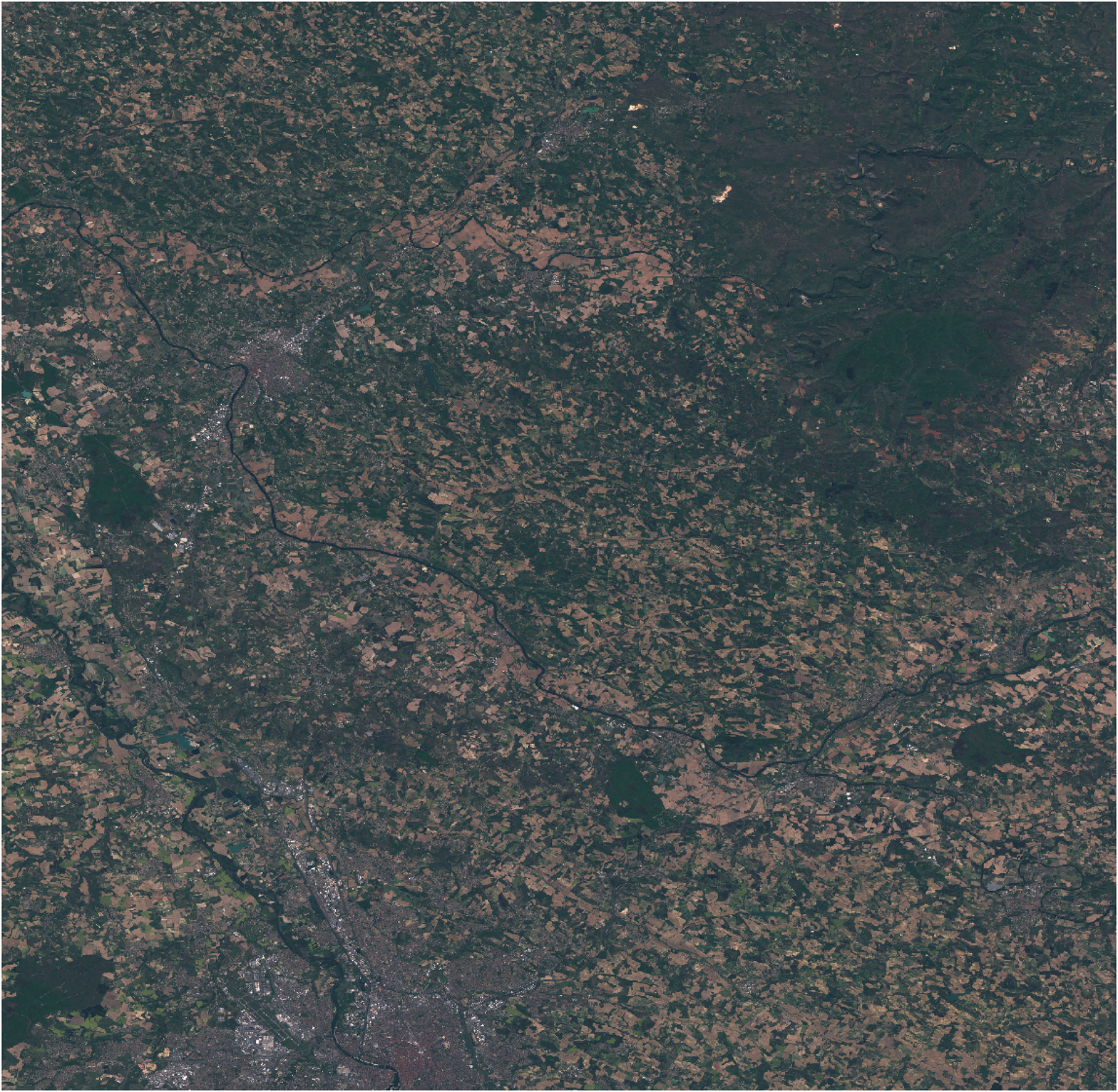}}
\hfill
\centering
\subfigure[February 2, 2020]{\includegraphics[width=6cm]{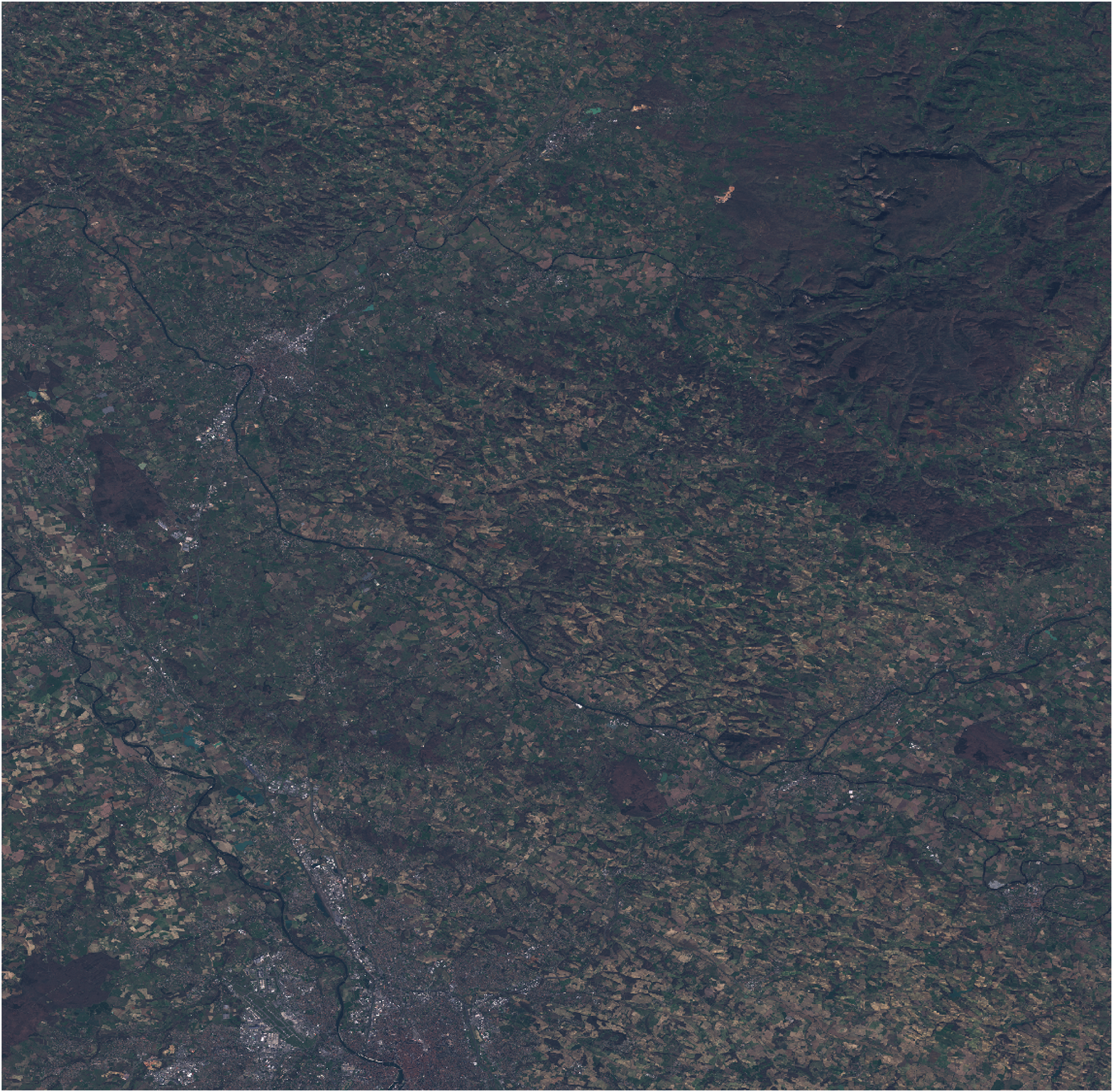}}
\hfill
\centering
\subfigure[January 11, 2020]{\includegraphics[width=6cm]{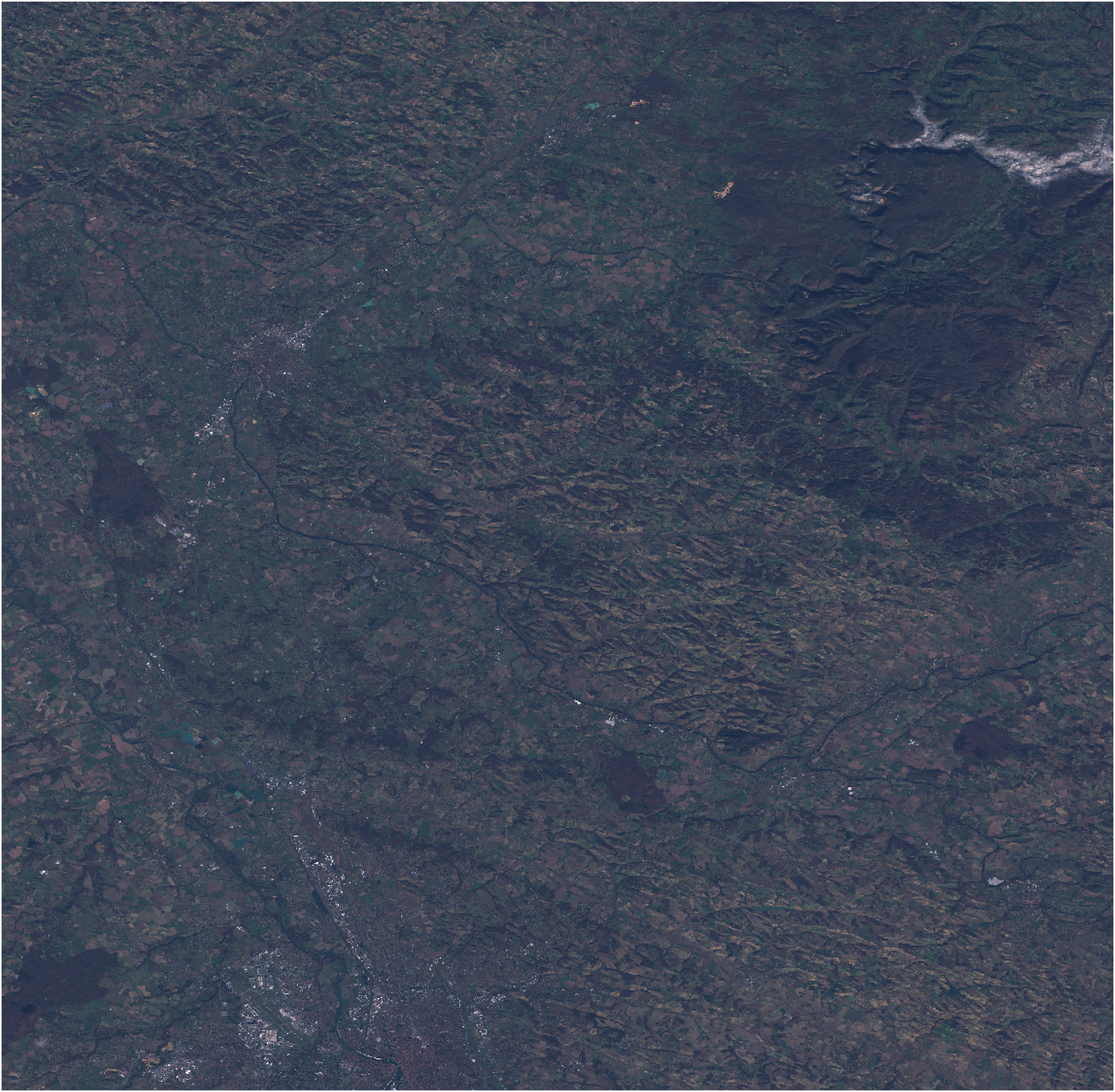}}
\caption{Sentinel-2 images of Toulouse taken on three different days in 2020.}
\label{day}%
\end{figure}

Both Sentinel-2A and 2B are now acquiring pictures of the Earth every five days at 10-m spatial resolution, with 3-7 days revisit frequency (M. Drusch et al., 2012). These images depend on the surface reflectance, which is defined as the difference of illumination and variation of the proportion of light reflected from the ground to the satellite sensor. For several reasons, this surface reflectance varies every day. One factor for example, is the sun movement that changes the sun-target-sensor geometry constantly. As a consequence, this effect causes an additional alteration of the radiometric data on pixels with the same land cover and similar structure (Vázquez-Jiménez et al., 2017), (S. A. Soenen, 2005). Therefore getting a land cover map by using the structure of Fig. \ref{unet_architecture} or other models of the state-of-the-art on a single image introduces systematically a variance that depends on the day when the image was acquired. Fig. \ref{day} shows three images acquired over the city of Toulouse, France on three different days. We can see that pixel radiometry values vary over time. \\

\begin{figure}[htp]
    \centering
    \includegraphics[width=7cm]{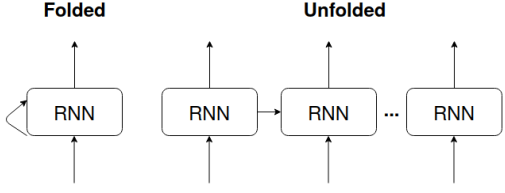}
    \caption{RNN unit (left) and unfolded structure (right)}
    \label{evolution}
\end{figure}

Time series images acquired by satellites contain highly-correlated information. Learning only relevant information from these time series and removing time-dependent variance can yield accurate and up-to-date land cover maps. To encode temporal dependencies in the PMap,  we can use RNNs. Differently from standard feed forward networks (e.g. FCN), RNNs explicitly manage temporal data dependencies since the output of the neuron at time $t-1$ is used, together with the next input, to feed the neuron itself at time $t$. A standard RNN unit is depicted on Fig. \ref{evolution}.

We focus on Long Short Term Memory (LSTM), an evolution of RNN which solves the problem of gradient explosion and gradient disappearance in RNNs. LSTM models were introduced by (Hochreiter, Schmidhuber, 1997) with the purpose to learn long term dependencies, since previous RNN models failed in this task. The input of our LSTM is a sequence of variables ($x_1$, ..., $x_N$) where a generic element $x_t$ is a PMap produced by the U-Net and $t$ refers to the corresponding day of image acquisition. RNN models are able to manage variable-length data sequences. \\

A standard LSTM unit is composed of a memory cell $c_t$, a hidden state $h_t$, and three different gates, the input gate $i_t$, the forget gate $f_t$, and the output gate $o_t$, that are employed to control the flow of information. All three gates combine the current input with the hidden state coming from the previous timestamp. The input gate $i_t$ decides how much of the current information enters the current memory cell while the forget gate $f_t$ decides how much information from the previous memory cell needs to be forgotten. Finally, the output gate $o_t$ decides how much information from the current memory cell $c_t$ will be outputted on the new hidden state $h_t$.  

\begin{figure}[htp]
    \centering
    \includegraphics[width=6cm]{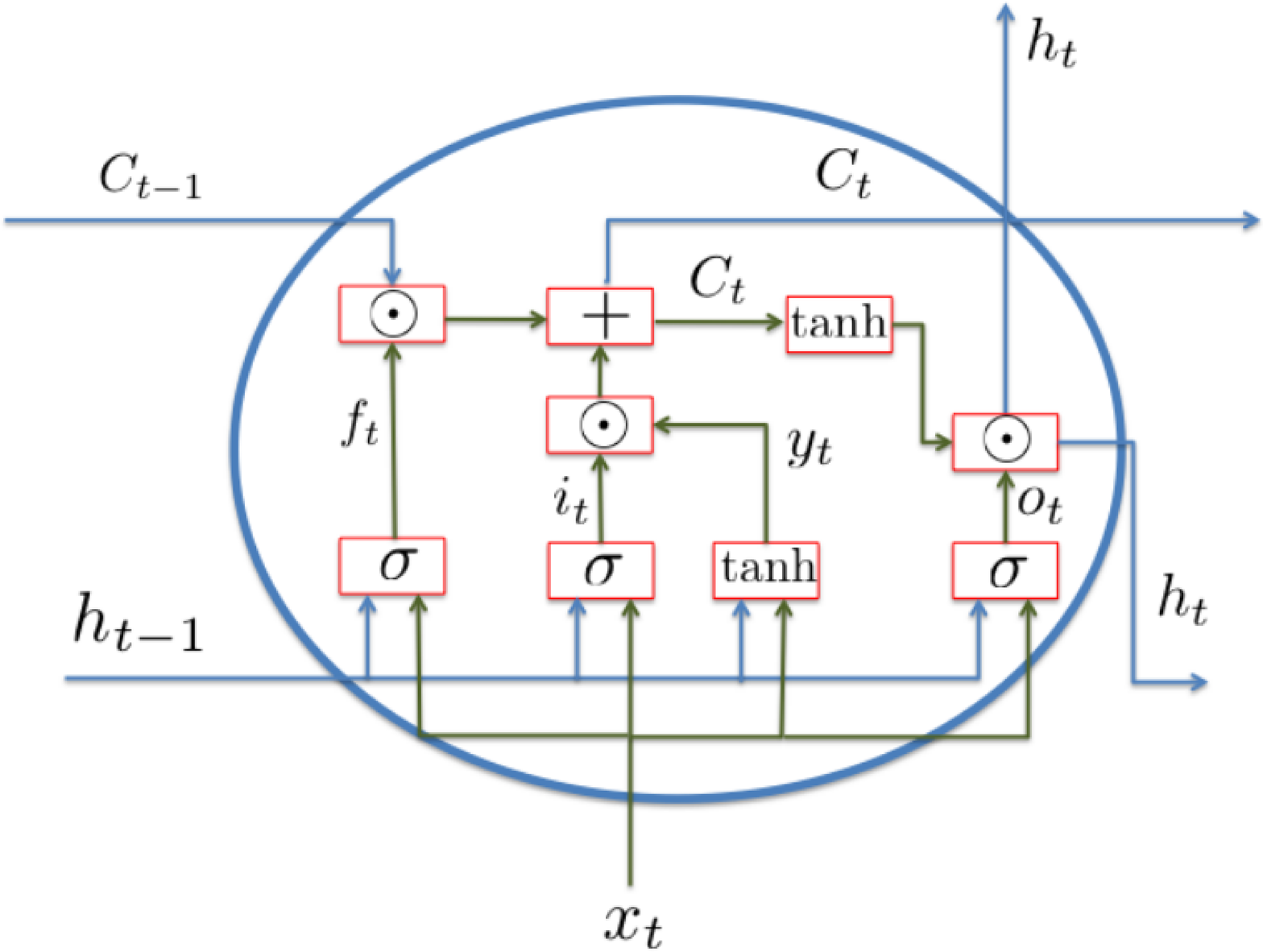}
    \caption{Internal structure of a typical LSTM unit. The arrows indicate directed connection, blue lines show the direction in which the information will flow while green lines underline internal flows. Red rectangles represent operations to combine or transform the different information.}
    \label{lstm}
\end{figure}

In this paper, LSTM is used to capture the correlation among the PMap generated by the FCN for the same land cover, but with a sequence of images taken at different dates. Then we aim to reduce the variance related to the day of acquisition and produce a \textit{single}, \textit{accurate} and \textit{up-to-date} probability map. When predicting subsequent attributes, LSTM can refer to the hidden state $h_t$ containing historical information. Although LSTM performs very well in sequence modeling tasks, the general LSTM ignores spatial information in an image during processing. This is because the standard LSTM models the sequence information through the full connection layer and flattens the input  image into a one-dimensional vector, which leads to loss of image spatial information. This is not optimal since features are correlated spatially within a single image and correlated temporally in a time series sequence. Retaining relevant spatial information is important for improving the performance. For the purpose of keeping the spatial structure of the feature map, we use Convolutional LSTM (ConvLSTM) where matrix multiplication is replaced by convolution at each gate. ConvLSTM networks capture spatiotemporal correlations better than standard fully connected LSTM. The key equations of ConvLSTM are shown in Equation (\ref{lstm}) below, where '$\star$' denotes the convolution operator and '$\cdot$' denotes the Hadamard product:

\begin{equation} 
\begin{split}
i_t & = \sigma \big( W_{xi} \star x_t  + W_{hi} \star h_{t-1} + W_{ci} \cdot c_{t-1} + b_i) \\
f_t & = \sigma \big( W_{xf} \star x_t  + W_{hf} \star h_{t-1} + W_{cf} \cdot c_{t-1} + b_f) \\
c_t & = f_t \cdot c_{t-1} + i_t \cdot tanh \big( W_{xc} \star x_t  + W_{hc} \star h_{t-1} +  b_c) \\
o_t & = \sigma \big( W_{xo} \star x_t  + W_{ho} \star h_{t-1} + W_{co} \cdot c_{t-1} + b_o) \\
h_t & = o_t \cdot \tanh(c_t)
\end{split}
\label{lstm}
\end{equation}
The weights $W$  of the connections in the LSTM are learned during the training and they determine how the gates operate. Each gate, referred by subscripts $f$, $i$, $o$, is controlled by trainable weights for the input, $W_{xi}$, $W_{xf}$, $W_{xo}$ $\in \mathbb{R}^{k\times k \times d \times r}$, for the hidden state, $W_{hi}$, $W_{hf}$, $W_{ho}$ $\in \mathbb{R}^{k\times k \times d \times r}$, and the memory cell $W_{ci}$, $W_{cf}$, $W_{co}$ $\in \mathbb{R}^{k\times k \times d \times r}$. The biases $b_i$ , $b_f$ , $b_c$ and $b_o$ $\in \mathbb{R} $ are also trainable parameters. $d$ is the input image channel number, $k$ is the convolutional kernel size, and $r$ is the hyperparameter that determines the number of filters in the recurrent layer. The sigmoid, $\sigma$, and the hyperbolic tangent $\tanh$, are used as activation functions in the gates. Since the input of the ConvLSTM is a PMap, $d = l_c = 2$. We set $k = 3$ and $r = 32$ filters.

\subsection{Overall model}
\begin{figure}[htp]
    \centering
    \includegraphics[width=9cm]{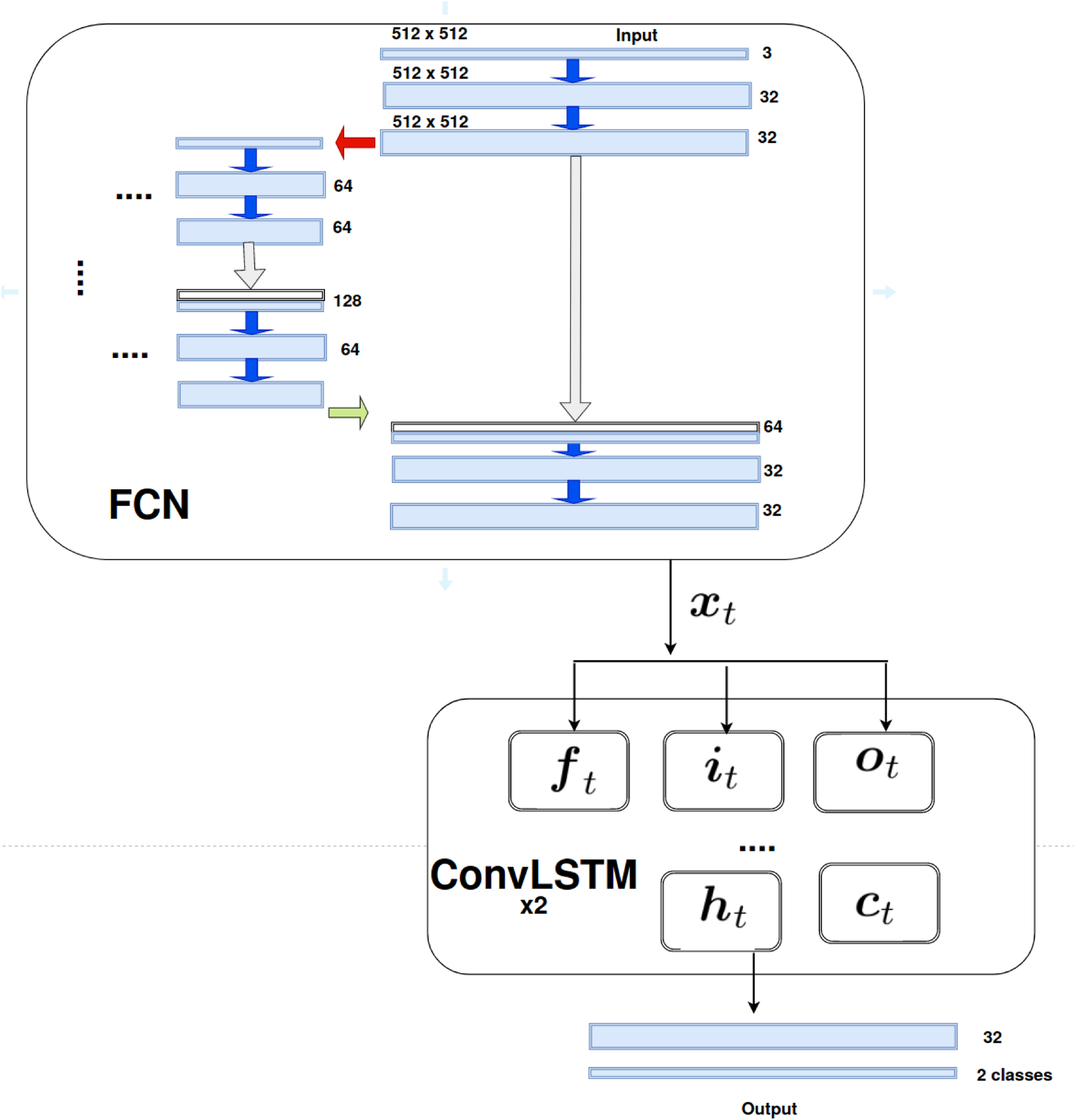}
    \caption{Proposed architecture for the model}
    \label{overall}
\end{figure}

The final architecture is shown on Fig. \ref{overall}.  It consists of a FCN, two ConvLSTM layers and one convolution layer. Gerenerally, two LSTM layers are enough to detect complex features. More layers can be better but also harder to train. The output of the FCN is sent to the first convLSTM layer. The input is a sequence of images from the same land cover but taken at different times such that we have enough variability in the surface reflectance. The images in the sequence are first encoded with the FCN to extract the probability maps. For each image, the output of this operation is an image of the same size as the original one but with $l_c$ channels (number of land cover classes). The first convLSTM combines this transformed image ($x_t$ in the equations) with its short-term $h_{t-1}$ and long-term $c_{t-1}$  memories. The output is another $l_c$-channels image that is sent to the second convLSTM layer. Finally, the short-term memory output, $h_t$ , is linearly transformed per pixel with the last convolution layer and scaled with a sigmoid function to obtain the final probability map.

We used a joint loss function $L$, combining cross entropy with a differentiable form of intersection over union (V. Iglovikov et al., 2017):

\begin{equation}
    L = \alpha H - (1-\alpha) \log(J),
\end{equation}
with $\alpha$ an hyperparameter that we can tune, $H$ the cross entropy and $J$ the IoU. 

\section{Dataset}
\label{dataset}
\begin{figure}[htp]
    \centering
    \includegraphics[width=9cm]{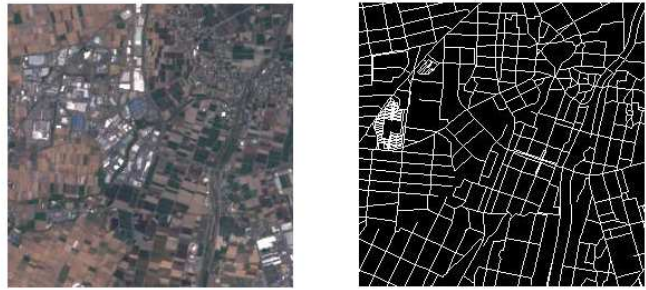}
    \caption{Image and its corresponding OSM reference data}
    \label{image_label}
\end{figure}

\begin{figure}[htp]
    \centering
    \includegraphics[width=6cm]{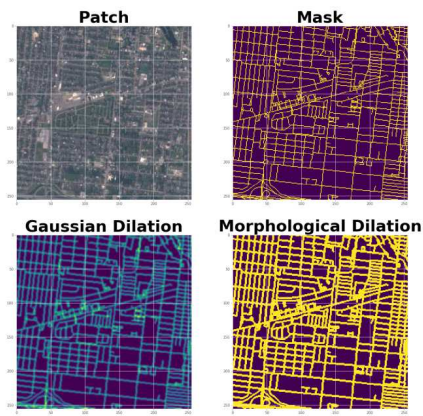}
    \caption{Dilation of road labels}
    \label{dilation}
\end{figure}

A key point of our study was to create a dataset. For that we needed to identify which geographic areas to include and which land cover
classes to consider. We focused  on regions where both the images and the reference data are available.
In addition, we required the data to be public and free. Given these constraints,  we used online map data from OpenStreetMap database which provides semantic labeling for many places in the World. We selected places where the data are well annotated. These data were generated by volunteers who labeled aerial or satellite images, or by national mapping agencies that donated their labeled data to make it available to a wider public. For each area, the database has historical files that contain the most recent updates in the OpenStreetMap data. We used these files to create reference data. For this study we considered two semantic classes: \textit{roads} and \textit{not roads}. For this we had to extract the road shapefiles from the OSM database. While there are other classes present (e.g: trees, buildings, rails), the \textit{roads} class is the only one that is consistent across large areas. For example \textit{Buildings} are usually not aligned with the structure or not represented at all. This makes it difficult to derive a pixelwise semantic labeling. The next step was to select a number of candidate areas for the dataset. For this purpose we visually inspected road labels accross several areas and assess whether the roads are properly aligned with the images. In some regions (e.g., Africa, Asia, South America), data were poorly annotated and therefore were excluded from the dataset. After extensive research, we found that certain cities in North America and Europe satisfy our needs. 16 cities were chosen in Canada, USA, France and Germany. The corresponding Sentinel-2 images were downloaded from ESA website. The download folder contains the envelope of all resolutions including 10m, 20m and 60m. We used to the True Color Image (TCI) at 10m resolution, built from the B02 (Blue), B03 (Green), and B04 (Red) bands. Fig. \ref{image_label} shows an example of a TCI image and its corresponding reference label obtained by rasterizing OSM shapefile. We set the sequence length of the RNN to 3. So for each city, 3 images corresponding to 3 different acquisition days were downloaded. Each image has a spatial resolution of 10m/pixel and a size of 10900 x 10900 pixels. Since images are very large, we cut them into small patches of size 512x512 pixels to fit them into GPU memory. We also used use vertical/horizontal flips and 0/90/180/270 degrees rotations to augment data. Data augmentation helps in building a strong model which is less dependent on input image orientation. This is very helpful for our model to generalize to different regions other than regions in training set. Roads are represented with a line in the shapefile, and very often this line is not properly located at the center of the road. Therefore we used morphological dilation to increase the width of the road after rasterization as illustrated on Fig. \ref{dilation}. We also tested dilation with a gaussian filter, but results were better with morphological dilation.

\section{Experiments}
\label{results}
Our pipeline is built with Tensorflow 1.13. To evaluate the performances of our model, we consider the accuracy metric,
defined as the percentage of correctly classified pixels. Table. \ref{results} summarizes our experimental results.
\subsection{Training}

To fit large images into GPU memory, we divided the input image into smaller patches 512x512x3 pixels. 13 cities were used for training and 3 for validation as illustrated in Table . \ref{cities}. First, the FCN is trained and then used to generate PMap sequences of length 3 to train the RNN. For both, we used the Adam optimization algorithm, with a base learning rate of 0.1 for 10 epochs, that we decreased to 0.01 for another 10 epochs and finally to 0.001 for the last 10 epochs. Each epoch consisted of 8 mini-batches. Training the FCN took 30 hours on a single GPU for the FCN and 20 hours for the RNN. Working with single images gave us an accuracy of 92.1\% during training and 91.1\% during validation. However, considering sequence of 3 images significantly increased the accuracy to 95.1\% during for training and  93.4 \% for  validation.

\subsection{Testing}
\begin{figure}[htp]
    \centering
    \includegraphics[width=6cm]{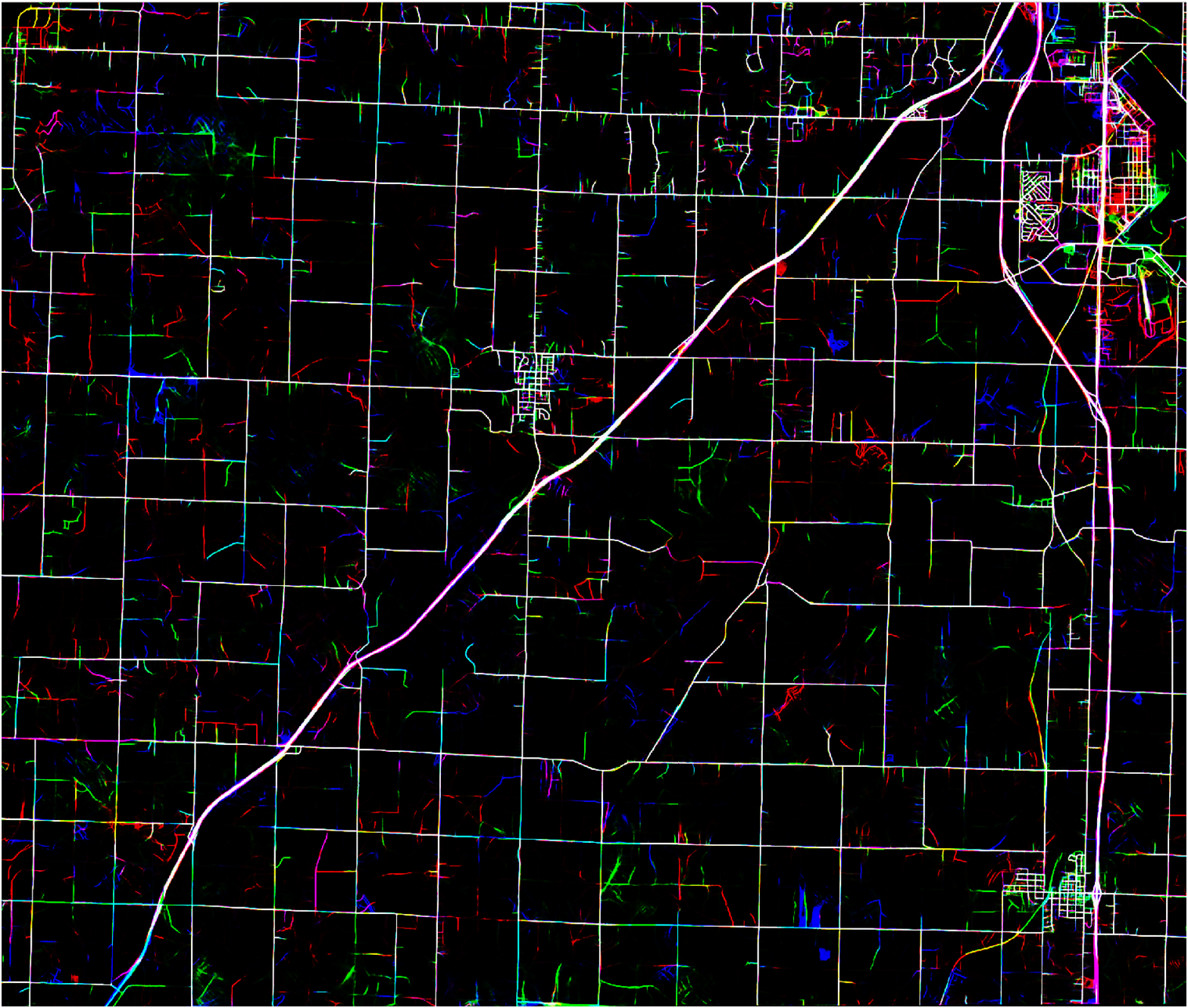}
    \caption{Road PMap generated by our FCN for the city of Topeka for three different days. Red is for March 11, green for July 4 and blue for October 7, 2019.}
    \label{PMap_unet}    
    \includegraphics[width=6cm]{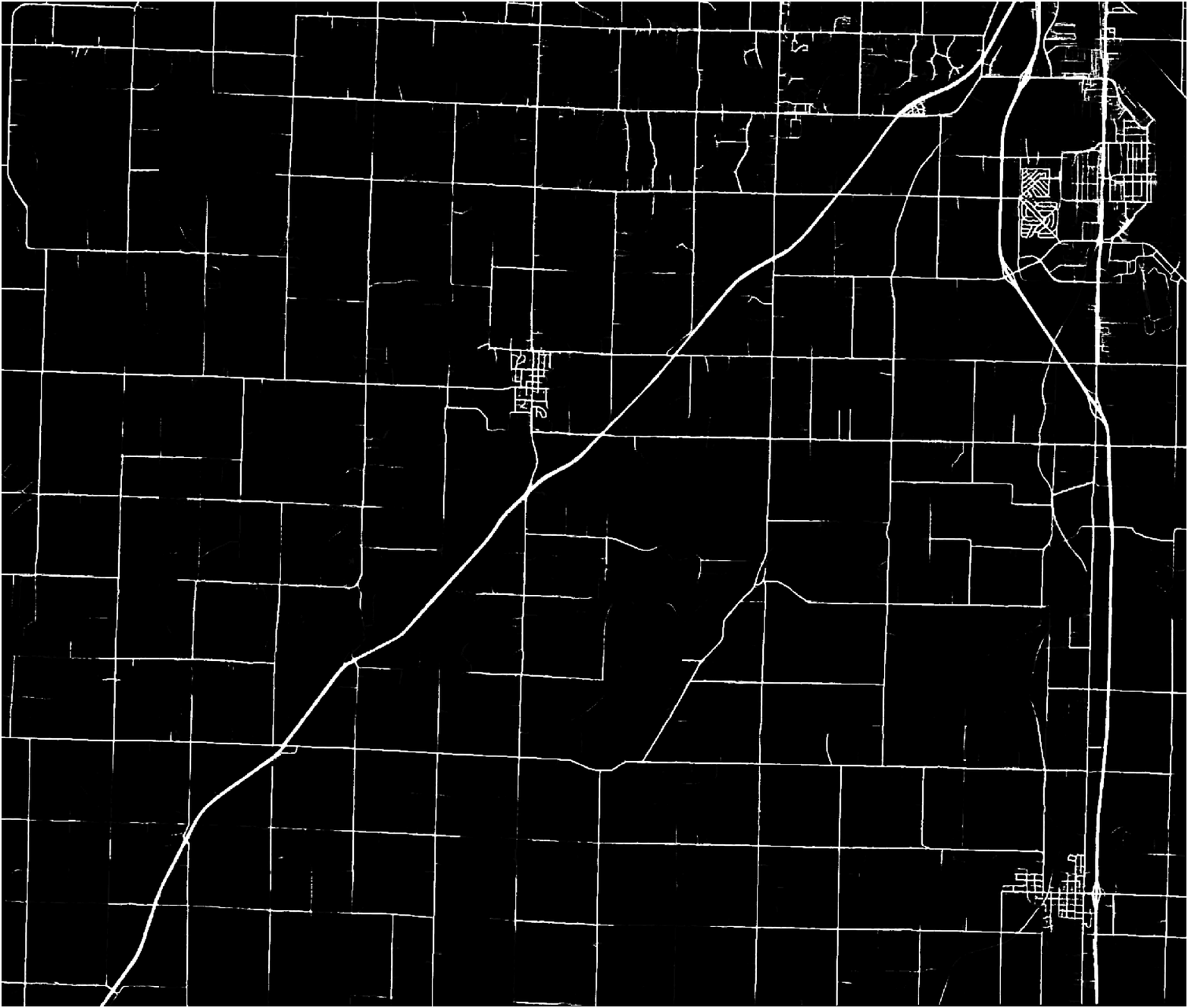}
    \caption{Road PMap generated by our ConvLSTM for the city of Topeka with a sequence of 3 images}
    \label{PMap_lstm}
    \includegraphics[width=6cm]{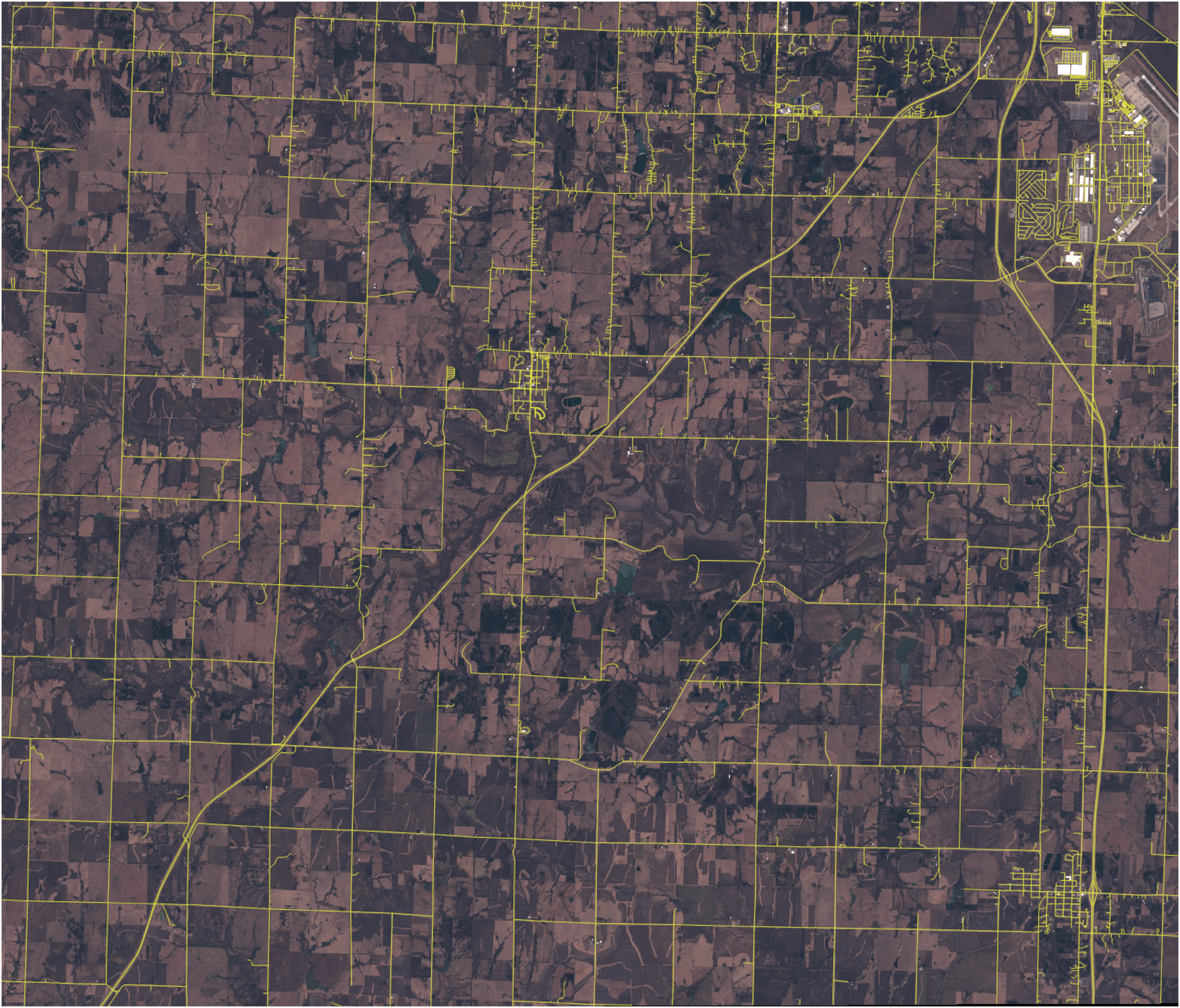}
    \caption{City of Topeka and groundtruth for road labels}
    \label{PMap_gt}
\end{figure}

To fit large images to GPU memory, we have cut them into patches of size 2048 $\times$ 2048 pixels (which was the maximum size that could be supported by our 2048 GPU), with 512 pixels of overlap between neighboring patches to avoid the border effect. Here we present testing results for the city of Topeka, USA. Fig. \ref{PMap_unet} shows the PMap obtained with the U-NET for the city of Topeka for three different days in 2019. The color red highlights the PMap obtained for the first day, the color green for the second day and the color blue for the third day. The color white for a given pixel indicates that all the three different PMaps predict a class \textit{roads} for that pixel. Mixed colors occur when two of the 3 predictions match. We can clearly see that there are some discrepancies among the 3 PMap, especially at some corners. In other words, the land cover mapping depends on the day when the image is acquired. This gives us an accuracy of 91.6\%. However, by combining the three different PMap with our RNN, we obtained the PMap of Fig. \ref{PMap_lstm}. By comparing this PMap with the ground truth of Fig. \ref{PMap_gt}, we can see that day-dependent variance is highly reduced and accuracy increases to  93.5\%. Producing accurate maps with this process can help us to track any change such as new roads.

\begin{table}[]
\begin{tabular}{|l|l|l|l|}
\hline
Model            & Training & Validation & Test \\ \hline
U-NET            & 92.1    & 91.1      & 91.6 \\ \hline
U-NET + ConvLSTM & 95.1     & 93.4       & 93.5 \\ \hline
\end{tabular}
\caption{Accuracy obtained with our models}
\label{results}
\end{table}

\begin{table}[]
\begin{tabular}{|l|l|l|l|}
\hline
\multicolumn{3}{|c|}{Training}       & Validation \\ \hline
Austin       & Dallas  & Edmonton    & Paris      \\ \hline
Jacksonville & Lyon    & Nantes      & Chesapeake \\ \hline
Houston      & Raleigh & Sacramento  & San Jose   \\ \hline
Dusseldorf   & Berlin  & San Antonio &            \\ \hline
Toronto      &         &             &            \\ \hline
\end{tabular}
\caption{Training and validation cities}
\label{cities}
\end{table}

\section{Conclusion}
In this paper, accurate and up-to-date land cover maps are generated by applying a deep learning model that exploits both temporal and spatial information provided by 10-m resolution multi-temporal and  multi-spectral satellite images. The deep learning model we designed  is based on the combination of a fully convolutional neural network with skip connections U-Net, which takes into account spatial information, together with a convolutional LSTM layer, which exploits the temporal information. The proposed methodology is used to identify road networks from time series of Sentinel-2 images. Experimental results show that encoding the temporal information from the image time series into the LSTM layer memory cells improves significantly the segmentation performance.

\end{document}